%% file: main.tex
\documentclass[letterpaper]{article} 
\usepackage{aaai2026}  
\usepackage{times}  
\usepackage{helvet}  
\usepackage{courier}  
\usepackage[hyphens]{url}  
\usepackage{graphicx} 
\usepackage{natbib}  
\usepackage{caption} 
\usepackage{algorithm}
\usepackage{algorithmic}

\usepackage{newfloat}
\usepackage{listings}
\DeclareCaptionStyle{ruled}{labelfont=normalfont,labelsep=colon,strut=off} 
\floatstyle{ruled}
\newfloat{listing}{tb}{lst}{}
\floatname{listing}{Listing}
\usepackage{amsmath}
\usepackage{dsfont} 

\usepackage{amsfonts}
\usepackage{booktabs}

\newcommand{\model}{GRAM}

\title{Generalizable Slum Detection from Satellite Imagery with Mixture-of-Experts} %

\author{
    Sumin Lee\thanks{Equal contribution to this work.}\textsuperscript{\rm 1,2},
    Sungwon Park\footnotemark[1]\textsuperscript{\rm 1,2},
    Jeasurk Yang\thanks{Co-corresponding authors.}\textsuperscript{\rm 1}, 
    Jihee Kim\footnotemark[2]\textsuperscript{\rm 2},
    Meeyoung Cha\footnotemark[2]\textsuperscript{\rm 1,2} 
}
\affiliations{
    \textsuperscript{\rm 1} Max Planck Institute for Security and Privacy (MPI-SP)\\
    \textsuperscript{\rm 2} Korea Advanced Institute of Science and Technology (KAIST) \\
}

\begin{document}
\maketitle

\input{0_abstract}

\begin{links}
    \link{Datasets}{https://github.com/DS4H-GIS/GRAM}
\end{links}

\input{1_introduction}

\input{2_related-works}

\input{3_dataset}
\input{4_methods}
\input{5_results}
\input{6_discussion}

\bibliography{aaai26}
\newpage
\newpage
\input{7_appendix}

\end{document}

%% file: 0_abstract.tex
\begin{abstract}
Satellite-based slum segmentation holds significant promise in generating global estimates of urban poverty. However, the morphological heterogeneity of informal settlements presents a major challenge, hindering the ability of models trained on specific regions to generalize effectively to unseen locations. To address this, we introduce a large-scale high-resolution dataset and propose \textbf{\model{}} ({G}eneralized {R}egion-{A}ware {M}ixture-of-Experts), a two-phase test-time adaptation framework that enables robust slum segmentation without requiring labeled data from target regions. We compile a million-scale satellite imagery dataset from 12 cities across four continents for source training. Using this dataset, the model employs a Mixture-of-Experts architecture to capture region-specific slum characteristics while learning universal features through a shared backbone. During adaptation, prediction consistency across experts filters out unreliable pseudo-labels, allowing the model to generalize effectively to previously unseen regions. \model{} outperforms state-of-the-art baselines in low-resource settings such as African cities, offering a scalable and label-efficient solution for global slum mapping and data-driven urban planning.
\end{abstract}

%% file: 1_introduction.tex
\section{Introduction}
In 2003, UN-Habitat introduced a globally standardized definition to describe deprived urban settlements---commonly referred to as \textit{slums}---marking a pivotal shift in how urban poverty was conceptualized and measured worldwide~\cite{united2003challenge}. This universal framework, formally adopted by the United Nations, was established to identify and quantify urban deprivation. It defines slums as settlements lacking one or more basic living conditions, such as durable housing, sufficient living space, access to safe water and sanitation, and secure tenure. This standardized definition enabled the production of comparable global estimates of slum populations, shaping major policy agendas like the Sustainable Development Goals (SDGs) and supporting cross-national research efforts~\cite{UNHABITAT_Data}.

\begin{figure}[t!]
\centering
\includegraphics[width=0.5\textwidth]{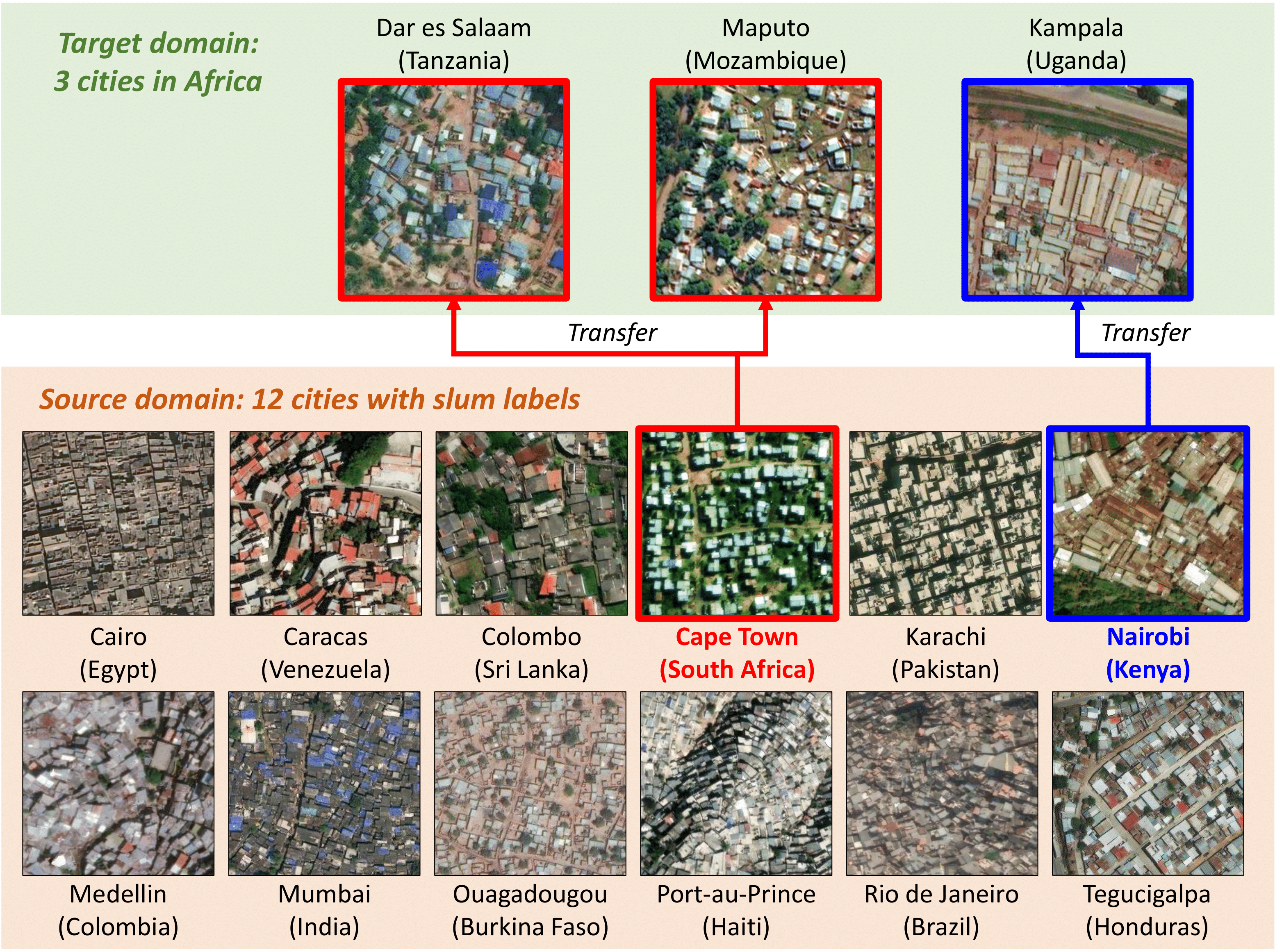}
\caption{Our source domain spans 12 cities across four continents; the target domain covers three African cities. Morphological similarities guide the model to prioritize Cape Town features for detecting slums in Dar es Salaam and Maputo (red), and Nairobi features for Kampala (blue).}
\label{fig: intro}
\end{figure}

Despite continued efforts, the worsening socioeconomic conditions in slums continue to present substantial challenges~\cite{kohli2012ontology}. Measuring these conditions is complex, as the physical form and living conditions vary widely—not only across continents but also within individual countries—reflecting differences in geography, governance, cultural norms, and historical trajectories of urbanization~\cite{taubenbock2014physical}. These localized differences make it difficult to apply standardized criteria for identification and comparison, thereby complicating conventional data collection methods~\cite{Yang2025Jurse}. A key limitation of traditional approaches to identifying slum settlements—such as surveys, censuses, and field assessments—is their variability in definitions and implementation. These methods differ across countries in how they conceptualize “slum,” the spatial scales they employ, and the degree to which political and institutional factors shape data collection. They are also resource-intensive to conduct. As a result, global estimates remain uneven, and cross-national comparisons are constrained by divergent definitions, methodologies, and practices on the ground~\cite{simon2011situating}.

Recent research has increasingly turned to satellite data and deep learning to measure urban deprivation~\cite{Kuffer2016}. High-resolution remote sensing data provide a globally accessible, frequently updated, and non-intrusive source of information, while convolutional neural networks and other computer vision techniques have shown promise in detecting spatial patterns associated with slum conditions~\cite{kit2013automated, Wurm2019slum, duque2017exploring}. Although promising, developing a computational approach is non-trivial due to visual heterogeneity~\cite{verma2019transfer, stark2020satellite}. Models trained on imagery from one country fail to generalize to other regions, as architectural styles, building materials, and spatial organization of slums differ (Figure~\ref{fig: intro}). As a result, cross-country and regional applications suffer from limited transferability and unreliable performance~\cite{Stark2024slum}.

To transfer knowledge from labeled source regions to unlabeled targets, we propose \textbf{\model{}}, a test-time adaptation (TTA) framework for slum segmentation. Unlike conventional methods that require annotated samples for targets, \model{} dynamically adapts a model pre-trained on source domains to the distributional characteristics of the test imagery for cross-regional generalization. To implement this idea, we construct a new slum segmentation dataset from very-high-resolution (VHR) satellite imagery of 12 cities in four continents, each representing distinct morphological, architectural, and socio-spatial patterns of informal settlements. These cities encompass diverse urban forms and regional contexts, offering a strong foundation for learning transferable representations. Using this multi-continent dataset and adaptation at inference time, we show how \model{} bridges the gap between universal slum definitions and region-specific visual characteristics.

The framework comprises two distinct training components. The first is the source model training phase, where Mixture-of-Experts (MoE) layers are integrated into the segmentation backbone. This architecture organizes the model into expert groups specialized in learning region-specific slum characteristics, while a shared backbone captures universal features across all regions. The second component is the target adaptation phase. For each unlabeled target image, a classifier identifies the source region with the most similar visual characteristics. The pseudo-label generated from that corresponding expert is used as a primary prediction. The model then assesses the image's reliability by evaluating the consistency of this reference against the predictions from all other experts. Images with the highest cross-expert agreement, measured by a ``stability score,'' are selected as reliable, effectively filtering out noise from uncertain pseudo-labels before the model is fine-tuned.

We validate \model{} using a diverse benchmark of slum segmentation tasks in three African cities with distinct urban morphologies and socio-spatial patterns. Our model consistently outperforms state-of-the-art baselines in low-resource settings, demonstrating its potential as a scalable and label-efficient solution for slum monitoring. These findings highlight the practical value of TTA for enabling more inclusive, data-driven urban development policies.

%% file: 2_related-works.tex
\section{Related Work}
\subsection{Satellite Data}
Satellite imagery provides a scalable and efficient alternative to traditional ground-based surveys of slums~\cite{ahn2023human,han2020lightweight,han2020learning}. Early approaches relied on proxy indicators, such as nighttime light intensity, to infer socioeconomic conditions from medium-resolution data~\cite{jean2016combining,park2022learning}. With advances in deep learning and VHR imagery, subsequent research shifted toward pixel-level segmentation.
For example, convolutional neural networks (CNNs) have been adapted for semantic segmentation of slum areas~\cite{lumban2023comparison}. Comparative studies have assessed various deep learning models in capturing the distinctive morphological characteristics of informal settlements, such as high-density and irregularly shaped structures~\cite{gadiraju2018machine, leoni2018machine}. Transfer learning~\cite{verma2019transfer, Wurm2019slum}, semi-supervised learning~\cite{rehman2022temporary, lin2024long}, and foundation model~\cite{zhang2024UVSAM} have also been applied, enabling models trained on large-scale datasets to adapt to slum segmentation tasks with limited labels.

In terms of data quality, both high- and medium-resolution imagery are being used for slum detection. One study explored the trade-offs between spatial detail and data accessibility~\cite{verma2019transfer}. Low-resolution multispectral data have also helped extend coverage in areas lacking VHR imagery~\cite{gramhansen2019mapping}. Additionally, machine learning--based slum mapping supports urban improvement efforts and enables monitoring of informal settlement dynamics over time~\cite{leoni2018machine, maiyababu2018slum,yang2025ai}.
Despite recent advances, segmenting slum settlements remains difficult due to significant variation in urban morphology across geographies. The scarcity of large-scale, pixel-level annotated datasets limits the development of fully supervised models. This constraint continues to drive research into developing new approaches including TTA that leverages abundant unlabeled satellite imagery to address data gaps.

\subsection{Test-Time Adaptation (TTA)}
Unsupervised domain adaptation improves model generalization by adapting a model trained on labeled data to a different, unlabeled domain with similar structure but differing data distribution. TTA builds on this idea under the stricter constraint that the source data is inaccessible during adaptation. Instead, TTA methods adapt a pre-trained source model \textit{during inference} using only the test sample data~\cite{ahn2025generalizable,liang2025comprehensive}.

Two widely used strategies include batch normalization (BN) adaptation and entropy minimization. In BN-based approaches~\cite{Nado2020EvaluatingPB}, the running statistics of the BN layers, originally computed during training, are updated using statistics from the target data, without requiring gradient computation~\cite{ioffe2015bn}. This provides a lightweight yet effective adaptation mechanism. Entropy-based methods aim to reduce the task uncertainty by minimizing the entropy of predictions on target samples. A typical approach is to fine-tune parameters, such as the affine components of BN layers, through backpropagation at test time~\cite{wang2021tent}.
While both strategies perform well on static domains, they face challenges in dynamic environments where target distributions shift over time or multiple domains are encountered sequentially. Under such conditions, performance tends to degrade due to accumulated errors and catastrophic forgetting. Recent work has explored continual adaptation to enhance model robustness across evolving and diverse target domains~\cite{wang2022continual,lee2024becotta}.


%% file: 3_dataset.tex
\section{Data}

For detecting slums, we construct a diverse labeled dataset of 12 cities listed below to cover a broad representation of urban morphologies and socio-economic conditions:
\begin{itemize}
\item \textbf{Africa (4 cities)}: Cairo in Egypt, Cape Town in South Africa, Nairobi in Kenya, Ouagadougou in Burkina Faso
\item \textbf{Asia (3 cities)}: Colombo in Sri Lanka, Karachi in Pakistan, and Mumbai in India
\item \textbf{South America (3 cities)}: Caracas in Venezuela, Medellín in Colombia, and Rio de Janeiro in Brazil
\item \textbf{Central America (2 cities)}: Port-au-Prince in Haiti, and Tegucigalpa in Honduras
\end{itemize}
Satellite imagery of these selected urban centers, sourced from the ESRI World Imagery Wayback\footnote{\url{https://livingatlas.arcgis.com/wayback/}}, is preprocessed into uniform 256$\times$256-pixel tiles at zoom level 16, corresponding to an approximate spatial resolution of 1.2 meters per pixel, subject to latitudinal variation.

\begin{figure}
\centering
\includegraphics[width=\columnwidth]{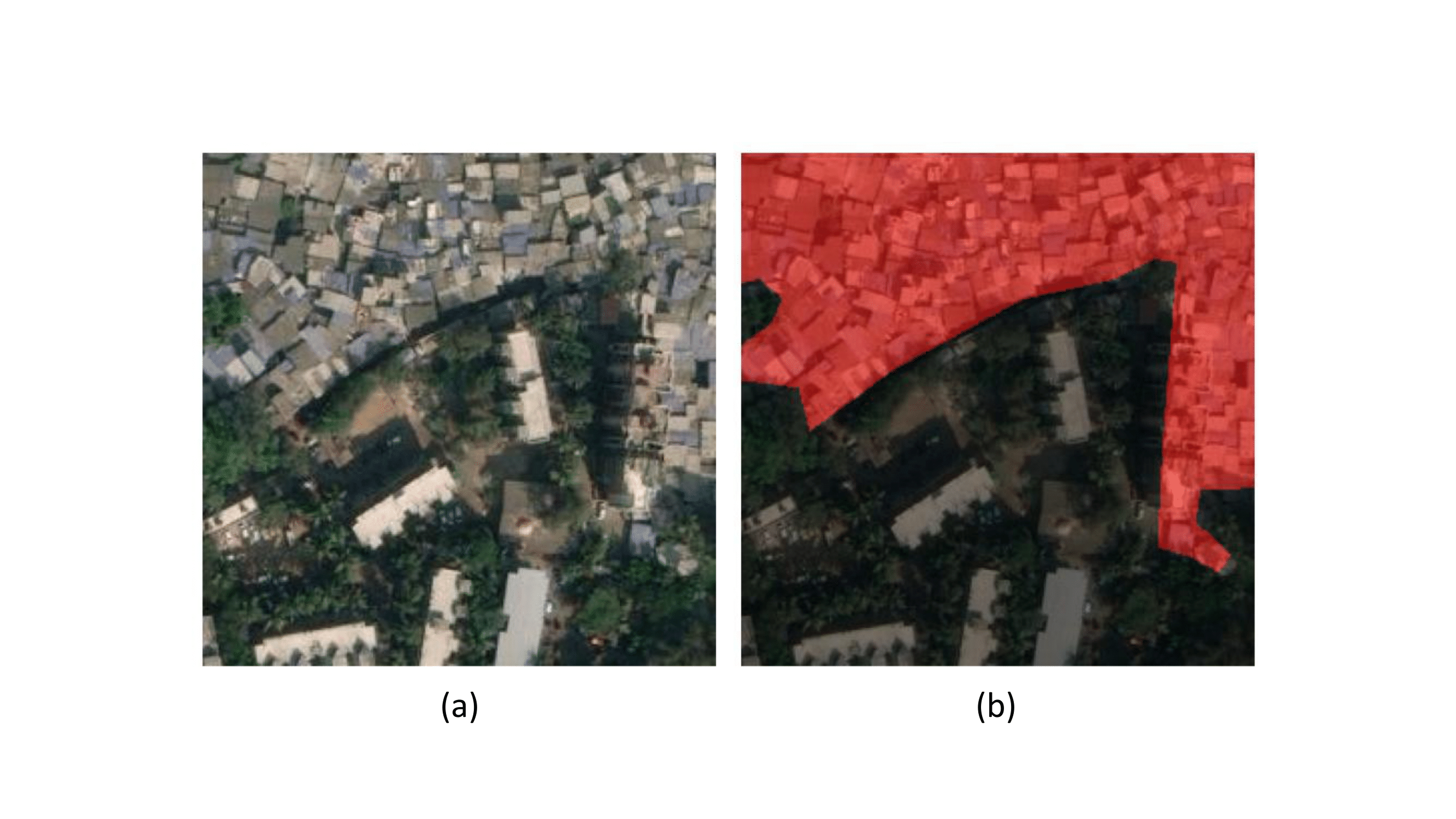}
\caption{Example of labeled satellite imagery used for training. (a) Original satellite image tile. (b) Corresponding binary mask overlaid in red, indicating slum areas (label=1), while black regions denote non-slum areas (label=0).}
\label{fig:label_example}
\end{figure}

Ground-truth annotations are from the Atlas of Informality~\cite{slumdata_AOI}, a global mapping initiative that documents the growth and transformation of informal settlements. We supplemented the labels with city-specific datasets~\cite{slumdata_CapeTown, slumdata_Karachi, slumdata_Mumbai, slumdata_Medellin, slumdata_RDJ}, each of which provides spatial delineations of informal settlements. Based on this resource, binary masks are manually generated by geography experts, where pixels corresponding to informal housing areas are assigned a value of 1, and all remaining pixels are set to 0. Each annotation is spatially aligned with its corresponding satellite tile, resulting in a 1$\times$256$\times$256 label image. An illustrative example is provided in Figure~\ref{fig:label_example}.

Using the manually labeled subset, we initially train a semi-supervised segmentation model based on the ST++ architecture~\cite{yang2022st_plus}. The model is trained on the full dataset comprising 2,714,489 image tiles, of which 86,752 have ground-truth annotations. This approach facilitates the integration of both labeled and unlabeled data to enhance segmentation performance under limited supervision. Subsequently, we use the model's predictions to generate pseudo-labels across the dataset. These pseudo-labels are then used to supervise the training of a fully supervised baseline model, enabling the use of a broader and more heterogeneous image set.

To test generalizability, we hold out a separate set of cities from the training process and use them exclusively for TTA and evaluation. This evaluation set comprises Dar es Salaam (Tanzania), Kampala (Uganda), and Maputo (Mozambique), totaling 529,633 image tiles, of which 17,536 are manually annotated using the AoI Database. The dataset details can be found in the Appendix.

To facilitate further research in slum detection and urban analysis, we release the dataset utilized in this study publicly available. This release includes both manually annotated segmentation labels and \model{}-derived masks across our 12 training and 3 testing cities. As of 2025, these 15 cities have an aggregated population of 120,174,837, with 215,148 informal settlement polygons identified by \model{}. This extensive coverage enables robust, cross-regional analyses of urban informality at scale. Each image tile has a spatial resolution of 10 meters per pixel at 256$\times$256 pixels, with corresponding geospatial coordinates aligned to the satellite imagery.

%% file: 4_methods.tex
\section{Method}
\textbf{Overview.} Let $\mathcal{D}_s$ denote the source training dataset and $\mathcal{D}_t$ the target test dataset. Each sample in the source dataset is a triplet $\{x, y, d\} \in \mathcal{D}_s$, where $x \in \mathbb{R}^{3 \times H \times W}$ is a satellite image, $y \in \{0,1\}^{H \times W}$ is the corresponding ground truth slum segmentation mask, and $d \in \{1, \ldots, D\}$ is the region label, with $D$ denoting the total number of source regions. Our goal is to develop a TTA framework that enables a source-trained model to generalize effectively to unseen target region images $x \in \mathcal{D}_t$.

Slum segmentation poses substantial difficulties due to the high variability in visual characteristics across geographies, cultures, and imaging conditions. This heterogeneity hampers the ability of a single model to generalize across domains. We propose \model{}, a two-stage training framework leveraging a MoE architecture to enhance adaptability and robustness in diverse environments, as outlined below:

\begin{figure*}[t!]
\centering
\includegraphics[width=1.0\textwidth]{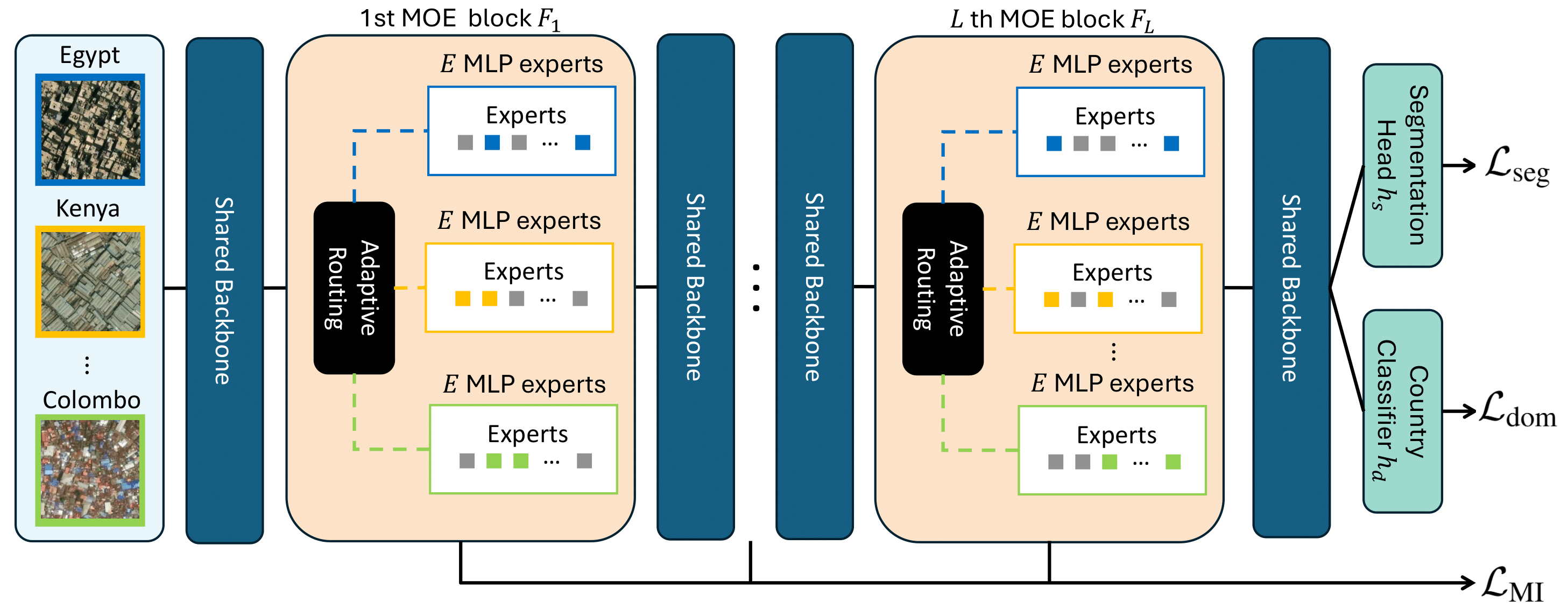}
\caption{Overview of the Mixture-of-Experts (MoE) architecture in \model{}. The diagram illustrates the integration of lightweight MoE blocks $\mathcal{F}$ into the transformer encoder, with region-specific gating networks $g_d$ dynamically routing token features $z$ to a top-$k$ subset of expert adapters $\{\mathcal{E}_e\}_{e=1}^E$.}
\label{fig:main_model}
\end{figure*}

\begin{itemize}
\item \textbf{Step 1.} Train the segmentation model $f_\theta$ using modular expert routing and region-aware learning on the multi-region slum dataset $\mathcal{D}_s$.
\item \textbf{Step 2.} Adapt the model $f_\theta$ on $ \mathcal{D}_t $ using pseudo-labels generated with routing guided by an external classifier $ h_\psi $, filtering out unreliable labels via cross-region prediction consistency, and fine-tuning on a high-stability subset $ \bar{\mathcal{D}}_t$. 
\end{itemize}

\subsection{Step 1. Source Training with Mixture-of-Experts}

While slums often share common morphological features---such as dense roof coverage and spatial layout---they also display region-specific variations in roof materials and construction styles. To disentangle these localized features from globally shared representations, we integrate lightweight Mixture-of-Experts (MoE) blocks $\mathcal{F}$ into $L$ intermediate layers of the transformer encoder, followed by a segmentation head $h_s$ (i.e., $f_\theta = h_s \circ \mathcal{F}_L \circ \cdots \circ \mathcal{F}_1$), as illustrated in Figure~\ref{fig:main_model}. Each MoE block $\mathcal{F}$ contains a set of lightweight MLP expert adapters $\{\mathcal{E}_e\}_{e=1}^{E}$, where $E$ is the number of experts. These experts capture region-specific slum characteristics, while the shared transformer backbone learns generalizable representations across diverse geographical contexts.

\paragraph{Adaptive Expert Routing within MoE Blocks} To enable region-specific specialization, we initialize a lightweight gating network $g_d$ for each source region $d$. Given a token feature $z$ extracted from an input image $x \in \mathcal{D}_s$, each MoE block dynamically selects a top-$k$ subset of experts for $z$ using a noisy top-$k$ routing strategy. The gating network for region $d$ computes logits $g_d(z) \in \mathbb{R}^E$, where each element indicates the relevance of an expert for processing $z$. To promote diversity and prevent overconfident expert selection~\cite{chen2023mod}, we add Gaussian noise to the logits:
\begin{align}
    \tilde{g}_d(z) &= g_d(z) + \epsilon, \quad \epsilon \sim \mathcal{N}(0, \sigma^2).
\end{align}

After injecting noise, we select the top-$k$ experts with the highest scores. Then, a softmax is applied over just these top-$k$ scores to produce a normalized set of weights  $\alpha$, which determine how much each selected expert contributes:
\begin{align}
    \boldsymbol{\alpha} &= \operatorname{Softmax}\left( \tilde{g}_d(z)_{\text{top-}k} \right), \\
    \text{MoE}(z) &= \sum_{e \in \text{top-}k(\tilde{g}_d(z))} \alpha_e \cdot \mathcal{E}_e(z),
\end{align}
\noindent where \( \mathcal{E}_e(z) \) is the output of expert \(e\)-th expert and $\alpha_e$ reflects how much that expert should contribute for token $z$. This routing mechanism allows the model to flexibly adapt to different region by selectively activating experts that are most relevant to the token’s region-specific context.

\paragraph{Region-Aware Regularization} In a MoE setup, the model learns to route tokens to different experts. Without proper regularization, experts may converge to similar behaviors, undermining specialization. To encourage diversity, we introduce a mutual information (MI)-based regularization term that promotes distinct expert selections across regions.

During training, we estimate the joint distribution $P^l(d,e)$ for $l$-th MoE layer, representing the frequency of expert $e$ being selected for samples from region $d$ in that layer. The mutual information between domains and experts for each layer is:

\begin{align}
I^l(d;e) &= \sum_{d=1}^D \sum_{e=1}^E P^l(d,e) \log \left( \frac{P^l(d,e)}{P^l(d)P^l(e)} \right).
\end{align}

By maximizing $I^l(d;e)$ for each layer, we encourage a strong dependence between the region and the selected experts, ensuring that different regions activate distinct sets of experts across the MoE blocks. This fosters expert specialization and prevents mode collapse, where experts learn redundant representations. To achieve this, we minimize the loss $\mathcal{L}_{MI} = - \sum_{L=1}^E I^l(d;e)$, which effectively maximizes the mutual information across all layers and promotes the desired domain-expert alignment.

Additionally, to complement the MI regularization and further promote expert specialization, we incorporate a lightweight region classifier $h_d$ into the training framework. This classifier operates on intermediate features extracted from the shared backbone, encouraging the model to learn domain-discriminative representations that enhance the dependence between cities and expert selections. It is trained to predict the region label using standard cross-entropy loss $\mathcal{H}$:

\begin{align}
\mathcal{L}_{\text{dom}} =\frac{1}{|\mathcal{D}_s|} \sum_{(x, d) \in \mathcal{D}_s} \mathcal{H}(d, h_d(x)),
\end{align}

\noindent where $h_d(x)$ denotes the predicted region probabilities for the sample $x$. This auxiliary supervision synergizes with the MI term by improving the quality of features used in routing, thereby preventing mode collapse and fostering region-specific expertise in the MoE layers.

\paragraph{Training Objectives} For segmentation supervision, we apply pixel-wise cross-entropy loss between the predicted segmentation map $f_\theta(x)$, and the ground truth $y$.  The segmentation loss over the entire source dataset  $\mathcal{D}_s$ is 
\begin{align}
\mathcal{L}_{\text{seg}} &= \frac{1}{|\mathcal{D}_s|} \sum_{(x, y, d) \in \mathcal{D}_s} \mathcal{H}_p(y, f_\theta(x, d)),
\end{align}
\noindent where the cross entropy loss $\mathcal{H}_p$ is averaged over all pixels and classes.
Finally, the overall training objective combines segmentation accuracy, expert diversity, and domain awareness:
\[
\mathcal{L}_{\text{total}} = \mathcal{L}_{\text{seg}} + \lambda_{\text{MI}} \cdot \mathcal{L}_{\text{MI}} + \lambda_{\text{dom}} \cdot \mathcal{L}_{\text{dom}}
\]
where \( \lambda_{\text{MI}} \) and \( \lambda_{\text{dom}} \) are hyperparameters controlling the influence of each regularization component.

\subsection{Step 2. Target Adaptation with Pseudo Label} Given the trained source model $f_\theta$, we perform TTA on the target dataset $\mathcal{D}_t$ by generating pseudo-labels for unlabeled target images.  
An external region classifier $h_\psi$ predicts the source region index most similar to the target region, which is used to guide the routing in $f_\theta$ predicts segmentation masks for samples in $\mathcal{D}_t$. The model is refined using self-training with these pseudo-labeled samples, minimizing a segmentation loss. However, the variability in slum characteristics introduces significant distribution shifts relative to the source data, potentially leading to unreliable pseudo-labels from the source model trained on a different distribution.  To address this, we
propose an image-level selection strategy that filters out the unreliable pseudo label via consistency of predictions across region-routed experts, followed by self-training to refine the model by minimizing a segmentation loss on the selected pseudo-labeled samples~\cite{park2021improving,yang2022st_plus}.

\paragraph{Adaptive Self-Training via Cross-Region Prediction Consistency} For a target sample $ x \in \mathcal{D}_t $, we first infer the target region index $ d_t $ using the external classifier $ h_\psi $. Using this index $ d_t $, we generate a pseudo-labeled dataset $ \bar{\mathcal{D}}_t $ as follows:

\begin{align}
d_t &= \arg\max_d [h_\psi(x)]_d, \\
\bar{\mathcal{D}}_t &= \{ (x, \bar{y}_{d_t}) \mid x \in \mathcal{D}_t, \bar{y}_{d_t} = \arg\max_{c} f_\theta(x, d_t)\},
\end{align}

\noindent where $ h_\psi(x) $ outputs the predicted probabilities for source region indices, and where $\bar{y}_d=\arg\max_{c} f_\theta(x, d) $ is the pseudo-mask generated by routing through region index $ d $.

Next, we assess the uncertainty of each image by evaluating the consistency of predictions across different region-specific routings. The stability score $ s $ for an image $ x $ is computed as the mean intersection-over-union (mIoU) between the pseudo-label $ \bar{y}_{d_t} $ generated with $ d_t $ and the pseudo-masks obtained from routing through all other source region indices $ d \in \{1, \dots, D\} \setminus \{d_t\} $. The stability score is defined as:
\begin{align}
s(x) &=  \sum_{d \neq d_t} \text{mIoU}(\bar{y}_{d_t}, \bar{y}_d).
\end{align}

Finally, we construct the adaptive target dataset $ \bar{\mathcal{D}}_t $ by selecting the most reliable images with the highest stability scores, constituting a fraction $ \rho_s $ of the target dataset (i.e., $ |\bar{\mathcal{D}}_t| = \rho_s \cdot |\mathcal{D}_t| $). The model is fine-tuned on $ \bar{\mathcal{D}}_t $ using self-training, minimizing a pixel-wise cross-entropy loss:
\begin{align}
\mathcal{L}_{\text{target}} &= \frac{1}{|\bar{\mathcal{D}}_t|} \sum_{(x, \bar{y}) \in \bar{\mathcal{D}}t} \mathcal{H}_p (\bar{y}, f_\theta(x, d_t)).
\end{align}

%% file: 5_results.tex
\section{Experiment}
We evaluate our model in terms of its generalizability for slum segmentation across domains, focusing on three African cities as unseen target regions.

\input{table/baseline}

\subsection{Performance Evaluation}
\paragraph{Implementation details} Adaptation is performed in a fully unsupervised setting, without access to any ground truth labels from the target domain.  We use the SegFormer~\cite{xie2021segformer} backbone and train all models under identical settings using SGD with a learning rate of 0.0001 and momentum of 0.99. In our experiments, we set $\rho_s = 0.5$, $E = 12$, and $k = 2$. Please refer to the Appendix for additional details on training details. We release the code of our model to facilitate broader adoption and greater impact within the research community: \url{https://github.com/DS4H-GIS/GRAM}

\paragraph{Result} We compare its performance against several state-of-the-art TTA methods. The compared baselines include: 
(1) \textsf{Vanilla Source}: A standard segmentation model without MoE; (2) \textsf{MoE Source}: An MoE-based model without TTA; (3) \textsf{SHOT}~\cite{liang2020shot}: Aligns target features to a frozen source classifier via information maximization and self-supervised learning; (4) \textsf{TENT}~\cite{wang2021tent}: Adapts the model during inference by minimizing the prediction entropy by updating the batch normalization statistics; (5) \textsf{CoTTA}~\cite{wang2022continual}: A mean-teacher-based approach that incorporates stochastic weight restoration to mitigate error accumulation over time; (6) \textsf{SAR}~\cite{niu2023towards}: Improves robustness by optimizing the sharpness of the entropy surface; (7) \textsf{BeCoTTA}~\cite{lee2024becotta}: Extends continual TTA with a MoE adapter architecture. We modify its target adaptation mechanism by incorporating our cross-region consistency sampling, which better leverages region-specific experts and filters unreliable pseudo-labels during target adaptation.

Table~\ref{tab:baseline} displays the performance with respect to mIoU, F1-score, precision, and recall.  Our model consistently outperforms all evaluated baselines across multiple cities. The superior performance of the MoE Source over the Vanilla Source demonstrates that the MoE architecture enhances generalizability during the source training process. Although BeCoTTA outperforms other baselines, it inadequately captures target prediction reliability. These results highlight our model's robustness and effectiveness in slum detection, particularly under severe cross-regional domain shifts. While conservative approaches (e.g., entropy regularization) suffice for limited shifts, they struggle in slum segmentation due to significant source-target divergences. In contrast, our self-training framework utilize pseudo-labels filtered by cross-expert consistency, enabling aggressive adaptation, noise robustness, and enhanced generalization in diverse urban contexts.

\input{table/ablation}
\paragraph{Ablation Study}  
To evaluate the contribution of each component within our framework, we conduct an ablation study on the Dar es Salaam region. Table~\ref{table:ablation} summarizes the average performance across slum and non-slum classes under different ablation settings. We evaluate five configurations: the first two examine modifications to source training, while the remaining three focus on the target adaptation phase.

Removing either the domain alignment loss $\mathcal{L}_{\text{dom}}$ or the mutual information loss $\mathcal{L}_{\text{MI}}$ results in a performance decline, confirming the necessity of both components. The impact is particularly pronounced for $\mathcal{L}_{\text{MI}}$, whose removal leads to a significant drop in mIoU (0.734) and F1-score (0.823), underscoring its central role in promoting expert consistency and preserving informative predictions.

We evaluate the impact of pseudo-label selection by comparing three strategies: no filtering, confidence-based filtering, and temporal consistency filtering. Omitting filtering results in moderate degradation, indicating that noisy pseudo-labels impair adaptation. Confidence-based filtering performs the worst (mIoU: 0.463), likely due to overconfidence under domain shift. Temporal consistency filtering enhances robustness by leveraging temporal agreement across source training checkpoints. (mIoU: 0.837, F1: 0.907). Our full method achieves the best performance (mIoU: 0.859, F1: 0.921), validating the effectiveness of combining mutual information regularization, domain alignment, and consistency-aware pseudo-labeling. See the Appendix for additional baseline comparisons and component analysis.

%% file: table/baseline.tex
\begin{table*}[t]
\centering
\setlength{\tabcolsep}{1mm}
\begin{tabular}{l|cccc|cccc|cccc}
\toprule
\textbf{Method} & \multicolumn{4}{c|}{\textbf{Dar es Salaam (Tanzania)}} & \multicolumn{4}{c|}{\textbf{Kampala (Uganda)}} & \multicolumn{4}{c}{\textbf{Maputo (Mozambique)}} \\
\cmidrule(lr){2-5} \cmidrule(lr){6-9} \cmidrule(lr){10-13}
& \textbf{mIoU} & \textbf{F1-score} & \textbf{Precision} & \textbf{Recall} & \textbf{mIoU} & \textbf{F1-score} & \textbf{Precision} & \textbf{Recall} & \textbf{mIoU} & \textbf{F1-score} & \textbf{Precision} & \textbf{Recall} \\
\midrule
Vanilla Source & 0.681 & 0.792 & 0.746 & 0.890 & 0.716 & 0.814 & 0.805 & 0.824 & 0.800 & 0.888 & 0.877 & 0.902 \\
MoE Source      & 0.806 & 0.885 & 0.895 & 0.876 & 0.800 & 0.881 & 0.831 & \textbf{0.956} & 0.900 & 0.947 & \textbf{0.942} & 0.953 \\
SHOT             & 0.712 & 0.813 & 0.847 & 0.786 & 0.713 & 0.810 & 0.846 & 0.782 & 0.813 & 0.895 & 0.890 & 0.901 \\
TENT             & 0.691 & 0.800 & 0.755 & 0.889 & 0.716 & 0.814 & 0.809 & 0.819 & 0.802 & 0.889 & 0.878 & 0.903 \\
CoTTA            & 0.762 & 0.853 & 0.857 & 0.850 & 0.821 & 0.900 & 0.894 & 0.908 & 0.821 & 0.900 & 0.895 & 0.905 \\
SAR              & 0.700 & 0.807 & 0.764 & 0.889 &  0.748& 0.843 & 0.829 & 0.859  & 0.804 & 0.890 & 0.881 & 0.900 \\
BeCoTTA          & 0.741 & 0.836 & 0.900 & 0.793 & 0.844 & 0.911 & \textbf{0.965} & 0.866 & 0.904 & 0.949 & 0.938 & 0.965 \\
\midrule
\model{}            & \textbf{0.859} & \textbf{0.921} & \textbf{0.911} & \textbf{0.931} & \textbf{0.870} & \textbf{0.927} & 0.932 & 0.921 & \textbf{0.907} & \textbf{0.951} & 0.939 & \textbf{0.966} \\
\bottomrule
\end{tabular}
\caption{Comparison of various metrics (mIoU, F1-score, Precision, Recall) across baseline and proposed methods in three regions: Dar es Salaam (Tanzania), Kampala (Uganda), and Maputo (Mozambique).}
\label{tab:baseline}
\end{table*}

%% file: table/ablation.tex
\begin{table}[t!]
\centering

\setlength{\tabcolsep}{1mm}
\begin{tabular}{lcccc}
\toprule
\textbf{Component} & \textbf{mIoU} & \textbf{F1-score} & \textbf{Precision} & \textbf{Recall} \\
\midrule
w/o $\mathcal{L}_{dom}$                & 0.836 & 0.906 & 0.876 & \textbf{0.944} \\
w/o $\mathcal{L}_{MI}$                & 0.734 & 0.823 & 0.898 & 0.786 \\
No Filtering            & 0.818 & 0.893 & 0.910 & 0.878 \\
Confidence Filtering    & 0.463 & 0.501 & 0.821 & 0.514 \\
Temporal Consistency  & 0.837 & 0.907 & 0.884 & 0.933 \\
\midrule
Full Component                    & \textbf{0.859} & \textbf{0.921} & \textbf{0.911} & 0.931 \\
\bottomrule
\end{tabular}

\caption{Average performance across slum and non-slum classes for various ablation settings on Dar es Salaam}
\label{table:ablation}

\end{table}

%% file: 6_discussion.tex
\section{Discussion}

\subsection{Region-Aware Routing Mirrors Visual Similarity}

\begin{figure}
    \centering
    \includegraphics[width=1.03\columnwidth]{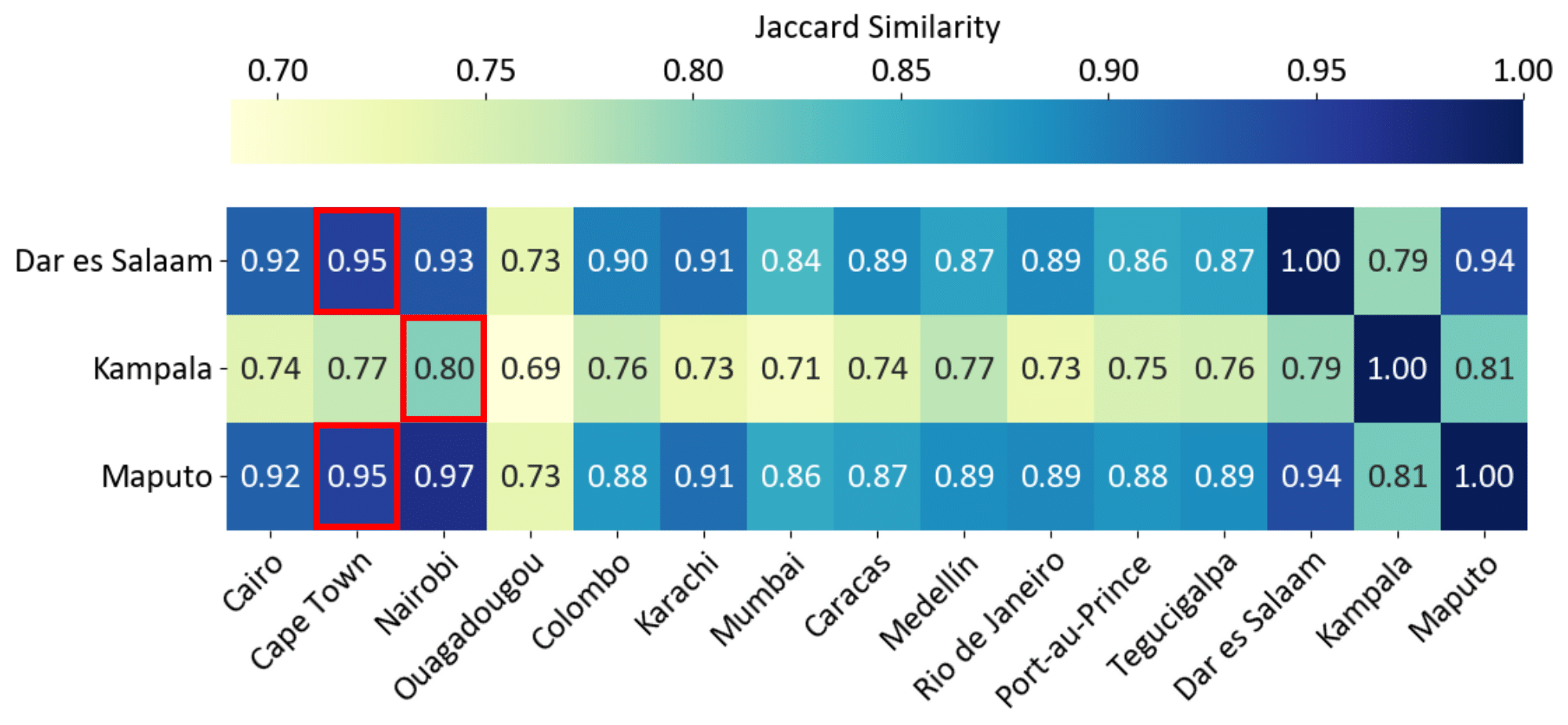}

    \caption{Jaccard similarity between the image sets of the three test cities (Dar es Salaam, Kampala, and Maputo) and those in the training set. Higher values indicate greater visual similarity in slum characteristics across cities. Red boxes denote the predictions made by the region classifier.}
    \label{fig:jaccard_matrix}
\end{figure}

Starting from the data complexity of capturing heterogeneous geographical patterns of slums, our method leverages the Mixture-of-Experts architecture for region-specific specialization in slum segmentation. During target adaptation, we utilize the output of an external region classifier $h_\psi$ to identify the source region most visually similar to the target region. Target samples are then dynamically routed through the corresponding expert set to facilitate adaptation.

To validate the effectiveness of this region-aware routing, we compute the Jaccard similarity between the image sets of source and target regions using visual features extracted from a pretrained CLIP model~\cite{Radford2021CLIP}. Specifically, we extract features from all images in the source and target datasets, apply K-means clustering in the feature space to assign discrete cluster labels, and calculate the Jaccard similarity between the label sets for regions $ A $ and $ B $. The results are visualized in Figure~\ref{fig:jaccard_matrix}. 

Region pairs predicted by the classifier (highlighted in red boxes---for example, Dar es Salaam and Cape Town) consistently exhibit higher similarity scores, indicated by the darker shades within the boxes. This correlation further supports the visual coherence of the classifier’s predictions. These matched pairs also reflect real-world geographic and socio-spatial patterns. For instance, Cape Town, Dar es Salaam, and Maputo are  coastal port cities, while Nairobi and Kampala are neighboring inland cities in East Africa. We can further associate the specific patterns of informal settlements with the examples shown in Figure~\ref{fig: intro}: where Cape Town, Dar es Salaam, and Maputo share square-like and light-gray features, whereas those in Nairobi and Kampala tend to be more rectangular and rusty brown. Importantly, the classifier is trained without explicit geographic information, suggesting that these associations are derived solely from the visual characteristics of slum regions.

\subsection{Slum Tracking Can Guide Policy}
\begin{figure}
    \centering
\includegraphics[width=1.00\columnwidth]{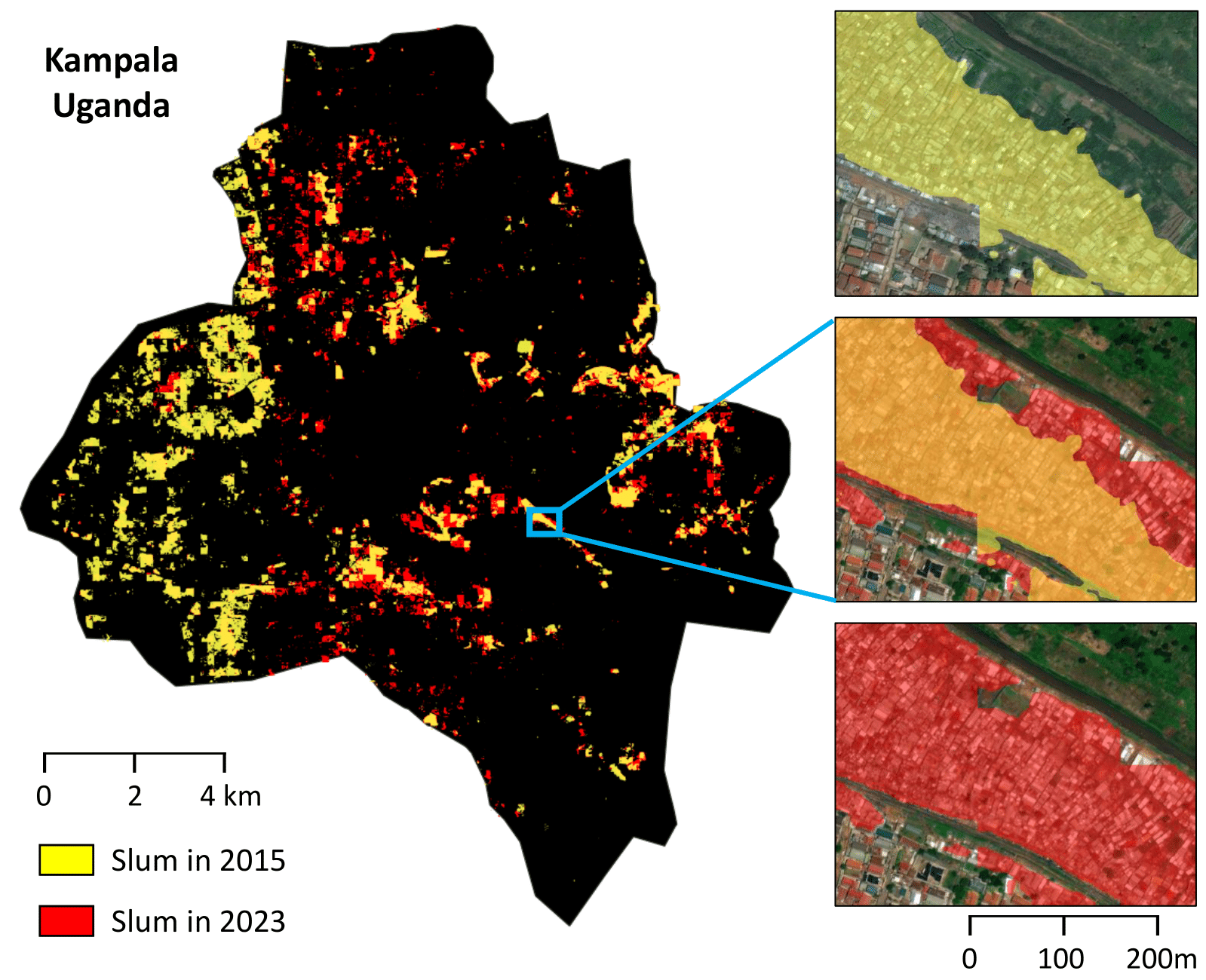}
    \caption{Slum segmentation results in Kampala in 2015 (yellow) and 2023 (red). Over the eight-year period, the slum ratio in the city increased from 8.4\% to 8.6\%.}
    \label{fig:inference}
\end{figure}

One of the goals of this work was to find a generalizable framework to help compute slum-related statistics for diverse regions where official data are scarce or unavailable. By applying our model to multi-temporal satellite imagery, stakeholders can now produce quantitative baseline estimates and monitor the evolution of informal settlements over time. For instance, our analysis revealed divergent trends in the three target cities: In Kampala, slum areas increased slightly from 8.4\% in 2015 to 8.6\% in 2023 (Figure~\ref{fig:inference}); Maputo experienced a sharp increase, from 35.3\% in 2016 to 41.2\% in 2023; while Dar es Salaam experienced a gradual decrease from 17.3\% in 2015 to 12.6\% in 2022.

These divergent outcomes are informative given that all three countries have experienced a comparable pace of economic growth. The ability to computationally track such nuanced, on-the-ground changes that broad economic indicators often overlook is tremendously valuable for urban policy making. Moreover, because the framework is label-efficient and required substantially little human effort compared to previous work, it can be scaled across a broader set of regions without requiring new, resource-intensive ground-truth data collection. 

This capability is a critical advantage, especially in low- and middle-income settings, where field surveys can be costly, politically sensitive, or logistically challenging.
The ability to accurately identify areas of greatest need allows for a more strategic and targeted allocation of resources, which is paramount for effective data-driven interventions. Furthermore, by publicly releasing a computation method to assess the long-term impact of urban policies, the framework ultimately empowers local decision-makers with the actionable evidence needed to address urban poverty, even in the absence of traditional census data.

\section{Conclusion}

We present \model{}, a TTA framework for slum detection in unseen regions, designed to tackle domain shifts in cross-regional settings. By integrating an MoE architecture with adaptive routing and cross-country prediction consistency, \model{} captures both shared and region-specific features while filtering unreliable pseudo-labels during self-training. Experiments on three African cities show that \model{} outperforms state-of-the-art baselines, demonstrating its effectiveness as a scalable, label-efficient solution for global slum monitoring. This approach holds promise in supporting inclusive, data-driven urban policy and can be extended to broader geographic contexts in future work.

\section*{Acknowledgments}
JKim was supported by the National Research Foundation of Korea (NRF) grant funded by the Korea government (MSIT) (No. RS-2022-NR068758 and RS-2025-00563196). We also thank the anonymous reviewers for their valuable comments and suggestions.

%% file: 7_appendix.tex
\appendix

\section{Appendix}
\label{sec:appendix}

\renewcommand{\thefigure}{S\arabic{figure}} 
\renewcommand{\thetable}{S\arabic{table}}   
\setcounter{figure}{0}
\setcounter{table}{0}

\subsection{Additional Results}

\input{table/appendix_TZA_result}

\begin{figure}[ht!]
    \centering
    \includegraphics[width=\linewidth]{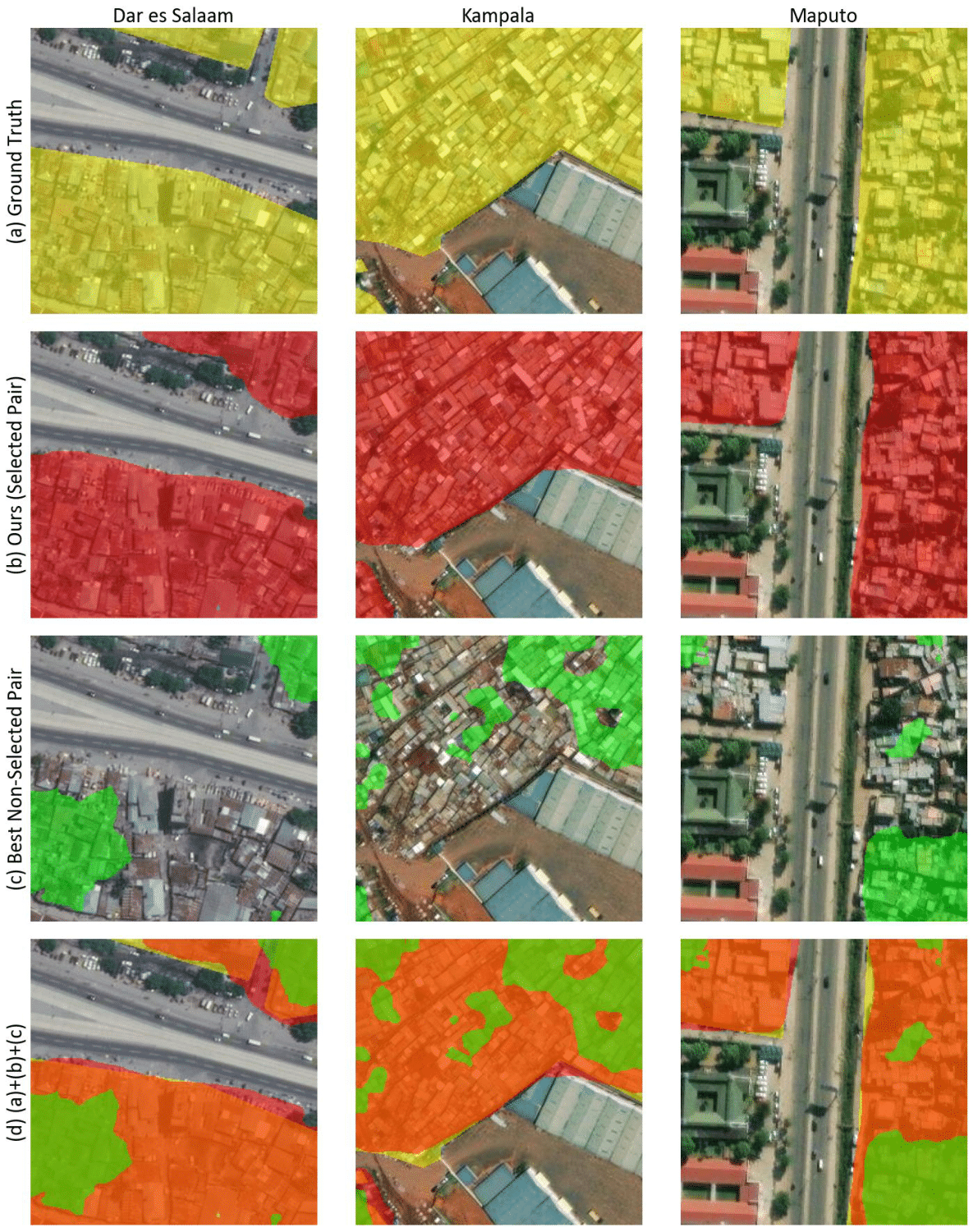}
    \caption{Visual comparison of segmentation results for Dar es Salaam (Tanzania), Kampala (Uganda), and Maputo (Mozambique) using different region pair configurations. (a) Ground truth (GT) segmentation maps. (b) Predictions generated using selected region pairs with external region classifier $h_\psi$: Dar es Salaam \& Cape Town, Kampala \& Nairobi, and Maputo \& Cape Town. (c) Predictions obtained from non-selected region pairs: Dar es Salaam \& Ouagadougou, Kampala \& Tegucigalpa, and Maputo \& Mumbai. (d) Overlay of (a), (b), and (c).}

    \label{fig:appendix_region_pair}
\end{figure}

\paragraph{Class-wise Results} Table~\ref{tab:full_results_TZA}  present the class-wise segmentation performance for each test region. As shown in the tables, \model{} consistently outperforms all baseline methods in both IoU and F1-score, with particular improvements in Dar es Salaam (with the increase in the IoU value from 0.659 to 0.752). While the MOE Source and BeCoTTA baselines achieve competitive results in terms of precision and recall, they fail to reach comparable IoU and F1-scores, which are metrics that better reflect overall segmentation performance and therefore suggest that \model{} is more robust under class-imbalanced conditions.

\paragraph{Comparison with Alternative Region Pairs}
To evaluate the effectiveness of our external region classifier $h_\psi$ and region-aware expert routing, we compare the performance of our selected region pairs against all alternative region pairs involving the same target city.

Across all three cities, the selected source regions consistently outperform alternative pairings. For Dar es Salaam, pairing with Cape Town yields a mean IoU of 0.859 and a slum IoU of 0.752, compared to 0.815 and 0.680 for the average of non-selected pairs. In Kampala, the selected source region, Nairobi, results in a mean IoU of 0.868 and a slum IoU of 0.759, again exceeding the corresponding averages (0.784 and 0.605). Maputo also benefits from pairing with Cape Town, achieving a mean IoU of 0.906 and a slum IoU of 0.873---surpassing the alternative pair average of 0.880 and 0.835, respectively.

\input{table/appendix_baseline}

These consistent improvements across all three cities suggest that our external classifier reliably identifies source domains that are better aligned with the target distribution. This targeted pairing allows \model{} to route inputs to more compatible experts, resulting in significantly improved segmentation, particularly in densely built-up slum areas. The strong results observed in three target regions demonstrate the robustness and generalizability of our region-pairing strategy. As illustrated in Figure~\ref{fig:appendix_region_pair}, this informed routing leads to more precise delineation of slum boundaries across diverse urban landscapes.

\paragraph{Additional Baselines} We conducted additional comparisons against previous slum and urban village detection baselines. The baselines include: (1) \textsf{TempSlum}~\cite{rehman2022temporary}, which expands seed labels through embedding-based pseudo-labeling; (2) \textsf{UV-SAM}~\cite{zhang2024UVSAM}, which adapts a vision foundation model via fine-tuning on urban village imagery; and (3) \textsf{LtCUV}~\cite{lin2024long}, which leverages a limited labeled dataset through a curriculum labeling strategy.
As shown in Table~\ref{tab:appendix_baseline}, our model outperforms these baselines by approximately 4.0\% in mIoU and 2.5\% in F1-score on average across the three test cities, demonstrating superior adaptability to diverse spatial morphologies and imaging conditions compared with prior approaches for informal or urban settlement mapping.

\paragraph{Dataset Contribution} Existing slum ground-truth datasets were often misaligned with satellite imagery due to geometric errors (Ref). We recalibrate these reference data to align with the RGB imagery while preserving original criteria (Aligned), with geographers manually re-labeling and achieving an inter-annotator mIoU of 0.84 for Mumbai. Using these manual labels, we trained a semi-supervised model to generate pseudo-labels for regions or periods lacking ground-truth, which were added to the source dataset. The table~\ref{tab:appendix_contribution} reports target mIoU from training SegFormer~\cite{xie2021segformer} on the source dataset without pseudo-labels (Ref and Aligned) and with pseudo-labels (+PL), demonstrating that our aligned and augmented dataset enhances transferability.

\input{table/appendix_contribution}

\paragraph{Dataset Splitting} To assess the temporal robustness of our approach, we divided the dataset according to acquisition time periods (see Implementation Details for the detailed timeline).
As shown in Table~\ref{tab:appendix_split}, our model maintained consistently stable performance, with a standard deviation of mIoU not exceeding 0.061 across all cities. On average, our model achieved mIoU of 0.871 and mF1 of 0.916, outperforming the vanilla source baseline (mIoU: 0.730, mF1: 0.763).
These results highlight the effectiveness of test-time adaptation (TTA) in mitigating temporal domain shifts in satellite imagery. Furthermore, the minimal performance fluctuation across time indicates that our model generalizes well to unseen temporal conditions, demonstrating its reliability for long-term monitoring and mapping tasks.

\input{table/appendix_split}

\paragraph{Ablation Details} To investigate the effect of the pseudo-label selection threshold $\rho_s$, we conduct an ablation study of Dar es Salaam (Tanzania) with different $\rho_s$ values from 0.1 to 0.5. Table~\ref{tab:rho_ablation} shows the result, where the best overall performance is achieved when $\rho_s=0.5$ with the highest IoU and F1 Score for both slum and non-slum classes. Although lower thresholds such as 0.4 and 0.3 lead to slightly higher precision and recall, they mark decreased performance for other classes. This exploration suggests that filtering pseudo-labels with moderate strictness, such as selecting the top 50\% most stable predictions, can effectively offer a balance between label reliability and coverage under class-imbalanced conditions.

\subsection{Implementation Details}

\input{table/appendix_rho}

\paragraph{Data Labeling}
\input{table/appendix_timeline}
We used satellite imagery from the ESRI World Imagery Wayback archive and utilized multi-temporal images to better support the model's generalizability. Table~\ref{tab:timeline} provides an overview of the satellite imagery acquisition dates for the 15 cities included in our study. Each column corresponds to a specific city, and each row represents a distinct acquisition timestamp. To capture seasonal and temporal variability, we selected images spanning multiple years and seasons, with most cities represented by five or more unique time points. This multi-temporal sampling strategy helps increase robustness to any visual changes in atmospheric conditions, vegetation, and lighting, thereby enhancing the generalizability of the segmentation model across diverse urban environments.

Table~\ref{tab:labeled_grid_numbers} summarizes the number of labeled and total grids for each city included in our dataset. The labeled grids correspond to image tiles annotated with slum and non-slum classes, sourced from publicly available datasets~\cite{slumdata_AOI, slumdata_CapeTown, slumdata_Karachi, slumdata_Mumbai, slumdata_Medellin, slumdata_RDJ}. The total grid count reflects the full set of $256 \times 256$ satellite image tiles at zoom level 16, used for model training and evaluation. While some cities contain relatively sparse labels (e.g., Kampala and Port-au-Prince), others, such as Cairo and Cape Town, offer extensive annotated coverage. This variability enables model performance under both high- and low-supervision regimes.

\newpage
\input{table/appendix_source}

\paragraph{Training} We use the SegFormer backbone, which is an efficient semantic segmentation model that offers several key advantages over traditional CNN-based and earlier transformer-based architectures, and train all models under identical settings using SGD with a learning rate of 0.01 for source training and 0.0001 for target adaptation and momentum of 0.99. The source model is trained for 3 epochs, followed by 10 additional epochs for target domain adaptation. Adaptation is performed in a fully unsupervised setting, without access to any ground truth labels from the target domain. In our experiments, we set $\rho_s = 0.5$, $E = 12$, and $k = 2$. Two NVIDIA GeForce RTX 3090 GPUs were used for each experiment. For baselines, we followed the hyperparameter configurations provided in the respective GitHub repositories.

%% file: table/appendix_TZA_result.tex
\begin{table}[ht]
\centering
\caption{Segmentation performance on Dar es Salaam (Tanzania),  Kampala (Uganda), and  Maputo (Mozam-
bique), reported separately for two classes across various metrics.}
\label{tab:full_results_TZA}
\resizebox{0.5\textwidth}{!}{
\begin{tabular}{l|cc|cc|cc|cc}
\toprule
\textbf{Dar es Salaam} & \multicolumn{2}{c|}{\textbf{IoU}} & \multicolumn{2}{c|}{\textbf{F1-score}} & \multicolumn{2}{c|}{\textbf{Precision}} & \multicolumn{2}{c}{\textbf{Recall}} \\
\cmidrule(lr){2-3} \cmidrule(lr){4-5} \cmidrule(lr){6-7} \cmidrule(lr){8-9}
 & Non-slum & Slum & Non-slum & Slum & Non-slum & Slum & Non-slum & Slum \\
\midrule
Vanilla Source & 0.885 & 0.476 & 0.939 & 0.645 & 0.984 & 0.508 & 0.898 & 0.882 \\
MOE Source     & 0.953 & 0.659 & 0.976 & 0.794 & 0.973 & 0.818 & 0.979 & 0.773 \\
SHOT             & 0.930 & 0.494 & 0.964 & 0.661 & 0.953 & 0.741 & 0.975 & 0.597 \\
TENT             & 0.893 & 0.489 & 0.943 & 0.657 & 0.983 & 0.527 & 0.907 & 0.871 \\
CoTTA            & 0.940 & 0.585 & 0.969 & 0.738 & 0.968 & 0.746 & 0.970 & 0.730 \\
SAR              & 0.899 & 0.502 & 0.947 & 0.668 & 0.983 & 0.545 & 0.914 & 0.864 \\
BeCoTTA          & 0.942 & 0.540 & 0.970 & 0.702 & 0.954 & \textbf{0.846} & \textbf{0.987} & 0.599 \\
\midrule
Ours             & \textbf{0.966} & \textbf{0.752} & \textbf{0.983} & \textbf{0.859} & \textbf{0.986} & 0.835 & 0.979 & \textbf{0.883} \\
\bottomrule
\end{tabular}
}
\resizebox{0.5\textwidth}{!}{
\begin{tabular}{l|cc|cc|cc|cc}
\toprule
\textbf{Kampala} & \multicolumn{2}{c|}{\textbf{IoU}} & \multicolumn{2}{c|}{\textbf{F1-score}} & \multicolumn{2}{c|}{\textbf{Precision}} & \multicolumn{2}{c}{\textbf{Recall}} \\
\cmidrule(lr){2-3} \cmidrule(lr){4-5} \cmidrule(lr){6-7} \cmidrule(lr){8-9}
 & Non-slum & Slum & Non-slum & Slum & Non-slum & Slum & Non-slum & Slum \\
\midrule
Vanilla Source & 0.942 & 0.490 & 0.970 & 0.658 & 0.973 & 0.637 & 0.967 & 0.680 \\
MOE Source     & 0.956 & 0.645 & 0.977 & 0.784 & \textbf{0.996} & 0.667 & 0.960 & \textbf{0.952} \\
SHOT             & 0.948 & 0.478 & 0.973 & 0.647 & 0.965 & 0.726 & 0.981 & 0.584 \\
TENT             & 0.943 & 0.490 & 0.970 & 0.657 & 0.972 & 0.646 & 0.969 & 0.669 \\
CoTTA            & 0.885 & 0.756 & 0.939 & 0.861 & 0.952 & 0.835 & 0.926 & 0.889 \\
SAR              & 0.932 & 0.564 & 0.965 & 0.721 & 0.971 & 0.687 & 0.959 & 0.759 \\
BeCoTTA          & 0.969 & 0.720 & 0.984 & 0.837 & 0.994 & \textbf{0.935} & 0.975 & 0.758 \\
\midrule
Ours             & \textbf{0.978} & \textbf{0.762} & \textbf{0.989} & \textbf{0.865} & 0.988 & 0.877 & \textbf{0.990} & 0.853 \\
\bottomrule
\end{tabular}
}
\resizebox{0.5\textwidth}{!}{
\begin{tabular}{l|cc|cc|cc|cc}
\toprule
\textbf{Maputo} & \multicolumn{2}{c|}{\textbf{IoU}} & \multicolumn{2}{c|}{\textbf{F1-score}} & \multicolumn{2}{c|}{\textbf{Precision}} & \multicolumn{2}{c}{\textbf{Recall}} \\
\cmidrule(lr){2-3} \cmidrule(lr){4-5} \cmidrule(lr){6-7} \cmidrule(lr){8-9}
 & Non-slum & Slum & Non-slum & Slum & Non-slum & Slum & Non-slum & Slum \\
\midrule
Vanilla Source & 0.868 & 0.733 & 0.930 & 0.846 & 0.955 & 0.798 & 0.905 & 0.899 \\
MOE Source      & 0.938 & 0.862 & 0.968 & 0.926 & 0.977 & 0.906 & \textbf{0.959} & 0.947 \\
SHOT             & 0.881 & 0.745 & 0.937 & 0.854 & 0.947 & 0.834 & 0.927 & 0.875 \\
TENT             & 0.870 & 0.735 & 0.930 & 0.847 & 0.955 & 0.802 & 0.907 & 0.898 \\
CoTTA            & 0.886 & 0.755 & 0.940 & 0.861 & 0.949 & 0.842 & 0.931 & 0.880 \\
SAR              & 0.873 & 0.735 & 0.932 & 0.847 & 0.950 & 0.813 & 0.915 & 0.885 \\
BeCoTTA          & 0.939 & 0.870 & 0.968 & 0.930 & 0.993 & 0.882 & 0.945 & 0.985 \\
\midrule
Ours             & \textbf{0.940} & \textbf{0.873} & \textbf{0.969} & \textbf{0.932} & \textbf{0.994} & \textbf{0.884} & 0.946 & \textbf{0.987} \\
\bottomrule
\end{tabular}
}
\end{table}

%% file: table/appendix_baseline.tex
\begin{table}[ht]
\centering
\caption{Performance comparison of baseline and proposed methods in Dar es Salaam  (Tanzania), Kampala (Uganda), and Maputo (Mozambique)}
\resizebox{0.5\textwidth}{!}{
\begin{tabular}{l|cc|cc|cc}
\toprule
\textbf{Method} & \multicolumn{2}{c|}{\textbf{Dar es Salaam}} & \multicolumn{2}{c|}{\textbf{Kampala}} & \multicolumn{2}{c}{\textbf{Maputo}} \\
 & mIoU & F1-score & mIoU & F1-score & mIoU & F1-score \\
\midrule
{TempSlum} & 0.818 & 0.894 & 0.857 & 0.919 & 0.872 & 0.931 \\
{UV-SAM} & 0.826 & 0.899 & 0.833 & 0.906 & 0.850 & 0.887 \\
{LtCUV} & 0.831 & 0.903 & 0.845 & 0.911 & 0.890 & 0.941 \\
\textbf{Ours} & \textbf{0.859} & \textbf{0.921} & \textbf{0.870} & \textbf{0.927} & \textbf{0.907} & \textbf{0.951} \\
\bottomrule
\end{tabular}}
\label{tab:appendix_baseline}
\end{table}

%% file: table/appendix_contribution.tex
\begin{table}[t!]
\centering
\caption{mIoU of SegFormer trained on source datasets with different labeling schemes: original misaligned ground truth (Ref), geometrically aligned and re-labeled data (Aligned), and our semi-supervised extension with pseudo-labels (+PL).}
\begin{tabular}{lccc}
\toprule
\textbf{Method} & \textbf{Dar es Salaam} & \textbf{Kampala} & \textbf{Maputo} \\
\midrule
Ref & 0.40 & 0.46 & 0.59 \\
Aligned & 0.45 & 0.54 & 0.66 \\
+PL & 0.68 & 0.72 & 0.80 \\
\bottomrule
\end{tabular}

\label{tab:appendix_contribution}
\end{table}

%% file: table/appendix_split.tex
\begin{table}[t!]
\centering
\caption{Temporal robustness evaluation of \model{} and vanilla source model across different time splits.}
\resizebox{0.5\textwidth}{!}{
\begin{tabular}{lcccc}
\toprule
\textbf{City} & \textbf{Metric} & \textbf{Ours} & \textbf{Vanilla Source} \\
\midrule
Dar es Salaam & mIoU & 0.842$\pm$0.050 & 0.679$\pm$0.048 \\
 & mF1 & 0.904$\pm$0.048 & 0.741$\pm$0.086 \\
 & mPrecision & 0.897$\pm$0.027 & 0.747$\pm$0.083 \\
Kampala & mIoU & 0.862$\pm$0.061 & 0.710$\pm$0.093 \\
 & mF1 & 0.899$\pm$0.007 & 0.676$\pm$0.217 \\
 & mPrecision & 0.882$\pm$0.012 & 0.730$\pm$0.158 \\
Maputo & mIoU & 0.909$\pm$0.015 & 0.800$\pm$0.020 \\
 & mF1 & 0.945$\pm$0.007 & 0.851$\pm$0.019 \\
 & mPrecision & 0.935$\pm$0.004 & 0.829$\pm$0.045 \\
\bottomrule
\end{tabular}}
\label{tab:appendix_split}
\end{table}

%% file: table/appendix_rho.tex
\begin{table}[t!]
\centering
\caption{Ablation study on the effect of the pseudo-label selection threshold $\rho_s$ in the Dar es Salaam (Tanzania) region. The results shows $\rho_s = 0.5$ yields the best  performance.}

\resizebox{0.5\textwidth}{!}{
\begin{tabular}{c|cc|cc|cc|cc}
\toprule
\textbf{$\rho_s$} & \multicolumn{2}{c|}{\textbf{IoU}} & \multicolumn{2}{c|}{\textbf{F1 Score}} & \multicolumn{2}{c|}{\textbf{Precision}} & \multicolumn{2}{c}{\textbf{Recall}} \\
 & non-slum & slum & non-slum & slum & non-slum & slum & non-slum & slum \\
\midrule
0.5 & \textbf{0.966} & \textbf{0.752} & \textbf{0.983} & \textbf{0.859} & 0.986 & 0.835 & 0.979 & 0.883 \\
0.4 & \textbf{0.966} & \textbf{0.752} & \textbf{0.983} & 0.858 & 0.985 & \textbf{0.840} & \textbf{0.980} & 0.878 \\
0.3 & 0.964 & 0.747 & 0.982 & 0.855 & \textbf{0.988} & 0.811 & 0.975 & \textbf{0.904} \\
0.2 & 0.958 & 0.713 & 0.978 & 0.833 & 0.987 & 0.780 & 0.970 & 0.893 \\
0.1 & 0.956 & 0.704 & 0.977 & 0.826 & 0.987 & 0.771 & 0.968 & 0.890 \\
\bottomrule
\end{tabular}
}

\label{tab:rho_ablation}
\end{table}

%% file: table/appendix_timeline.tex
\begin{table}[t!]
\centering
\caption{Overview of satellite imagery acquisition dates across 15 cities. Each column corresponds to a specific city, and each row represents a unique acquisition timestamp.}
\resizebox{0.5\textwidth}{!}{%
\begin{tabular}{llllllll}
\toprule
\textbf{Cairo} & \textbf{Cape Town} & \textbf{Nairobi} & \textbf{Ouagadougou} & \textbf{Colombo} & \textbf{Karachi} & \textbf{Mumbai} & \textbf{Caracas} \\
\midrule
Jul 2015 & Oct 2015 & Jan 2014 & Dec 2019 & Dec 2015 & Nov 2014 & Jan 2016 & Mar 2013 \\
Jan 2019 & Nov 2016 & Jan 2017 & Jan 2021 & Dec 2016 & Jan 2017 & Dec 2016 & Mar 2014 \\
Jul 2020 & Jan 2017 & Feb 2019 & Jan 2022 & Jan 2020 & Oct 2018 & Oct 2018 & Apr 2017 \\
Nov 2021 & Jan 2019 & Feb 2020 & Nov 2022 & May 2022 & Nov 2020 & Jan 2020 & Mar 2018 \\
May 2022 & Jan 2022 & Aug 2021 & Jan 2024 & Mar 2023 & Sep 2021 & Oct 2021 & Jan 2019 \\
Mar 2023 & Jan 2023 & Feb 2023 &         &         & Mar 2022 & Feb 2022 & Apr 2020 \\
Jul 2023 & Jan 2024 &         &         &         & Jan 2023 & Jan 2023 & Dec 2021 \\
        &          &         &         &         & Nov 2023 &         & Mar 2024 \\
\bottomrule
\end{tabular}

}

\resizebox{0.5\textwidth}{!}{%
\begin{tabular}{lllllll}
\toprule
\textbf{Medellín} & \textbf{Rio de Janeiro} & \textbf{Port-au-Prince} & \textbf{Tegucigalpa} & \textbf{Dar es Salaam} & \textbf{Kampala} & \textbf{Maputo} \\
\midrule
Jan 2015 & Jul 2017 & Jan 2016 & Dec 2015 & Aug 2018 & Feb 2015 & Jul 2016 \\
Feb 2017 & Jun 2018 & Dec 2016 & Dec 2016 & Mar 2017 & Nov 2016 & Apr 2017 \\
Aug 2019 & Oct 2019 & Dec 2017 & Mar 2018 & Mar 2018 & Jan 2018 & Dec 2018 \\
Jan 2020 & Sep 2021 & Nov 2018 & Mar 2019 & May 2021 & Dec 2019 & Mar 2020 \\
Nov 2020 & May 2022 & Mar 2020 & Apr 2020 & Jun 2022 & Jul 2021 & Jul 2021 \\
Aug 2023 & May 2023 & Jan 2021 & Mar 2022 & Feb 2023 & Feb 2023 & Oct 2022 \\
Jan 2024 &         & Jul 2021 & Jan 2023 &         &         & May 2023 \\
        &          & Mar 2023 &         &         &         & Apr 2024 \\
        &          & Jan 2024 &         &         &         &         \\
\bottomrule
\end{tabular}
}
\label{tab:timeline}
\end{table}

%% file: table/appendix_source.tex
\begin{table}[ht]
\centering
\caption{Summary of labeled and total grid counts across all cities included in the dataset.}
\resizebox{0.5\textwidth}{!}{
\begin{tabular}{lrr}
\toprule
\textbf{City} & \textbf{Labeled Grids} & \textbf{Total Grids} \\
\midrule
Cairo~\cite{slumdata_AOI}             & 11,174 & 393,158 \\
Cape Town~\cite{slumdata_CapeTown}         & 10,808 & 456,971 \\
Nairobi~\cite{slumdata_AOI}           & 6,505  & 122,764 \\
Ouagadougou~\cite{slumdata_AOI}       & 8,449  & 123,204 \\
Colombo~\cite{slumdata_AOI}           & 4,376  & 80,300  \\
Karachi~\cite{slumdata_Karachi}           & 11,612 & 319,328 \\
Mumbai~\cite{slumdata_Mumbai}            & 8,316  & 169,407 \\
Caracas~\cite{slumdata_AOI}           & 6,853  & 97,232  \\
Medellin~\cite{slumdata_Medellin}          & 3,950  & 83,256  \\
Rio de Janeiro~\cite{slumdata_RDJ}    & 9,179  & 353,496 \\
Port-au-Prince~\cite{slumdata_AOI}    & 2,344  & 29,286  \\
Tegucigalpa~\cite{slumdata_AOI}       & 3,528  & 110,988 \\
Dar es Salaam~\cite{slumdata_AOI}     & 6,050  & 237,974 \\
Kampala~\cite{slumdata_AOI}           & 2,623  & 48,696  \\
Maputo~\cite{slumdata_AOI}            & 8,863  & 242,963 \\
\bottomrule
\end{tabular}
}
\label{tab:labeled_grid_numbers}
\end{table}

%% file: aaai26.bib
@article{Stark2024slum,
  author		= "Stark, Thomas and Wurm, Michael and Zhu, Xiao Xiang and Taubenböck, Hannes",
  title			= "Quantifying uncertainty in slum detection: advancing transfer-learning with limited data in noisy urban environments",
  journal		= "IEEE Journal of Selected Topics in Applied Earth Observations and Remote Sensing.",
  volume		= "17",
  pages			= "4552--4565",
  year			= "2024"
}

@inproceedings{ioffe2015bn,
author = {Ioffe, Sergey and Szegedy, Christian},
title = {Batch normalization: accelerating deep network training by reducing internal covariate shift},
year = {2015},
booktitle = {proc. of the 32nd International Conference on International Conference on Machine Learning},
pages = {448--456},
numpages = {9},
}

@article{Kuffer2016,
  author		= "Kuffer, Monika and Pfeffer, Karin and Sliuzas, Richard",
  title			= "Slums from Space—15 Years of Slum Mapping Using Remote Sensing",
  journal		= "Remote Sensing",
  volume		= "8",
  number		= "6",
  pages			= "455",
  year			= "2016"
}

@article{jean2016combining,
author = {Jean, Neal and Burke, Marshall and Xie, Michael and Davis, W. and Lobell, David and Ermon, Stefano},
title = {Combining satellite imagery and machine learning to predict poverty},
journal = {Science},
volume = {353},
number = {6301},
pages = {790-794},
year = {2016},
}

@Article{lumban2023comparison,
AUTHOR = {Lumban-Gaol, Yustisi A. and Rizaldy, Aldino and Murtiyoso Amadi},
TITLE = {Comparison of deep learning architectures for the semantic segmentation of slum areas from satellite images},
JOURNAL = {The International Archives of the Photogrammetry, Remote Sensing and Spatial Information Sciences},
VOLUME = {XLVIII-1/W2-2023},
YEAR = {2023},
PAGES = {1439--1444},
}

@INPROCEEDINGS{gadiraju2018machine,
  author={Gadiraju, Krishna Karthik and Vatsavai, Ranga Raju and Kaza, Nikhil and Wibbels, Eric and Krishna, Anirudh},
  booktitle={proc. of the 2018 IEEE International Conference on Data Mining Workshops}, 
  title={Machine learning approaches for slum detection using very high resolution satellite images}, 
  year={2018},
  pages={1397-1404},
}

@Article{leoni2018machine,
AUTHOR = {Leonita, Gina and Kuffer, Monika and Sliuzas, Richard and Persello, Claudio},
TITLE = {Machine learning-based slum mapping in support of slum upgrading Programs: The case of Bandung City, Indonesia},
JOURNAL = {Remote Sensing},
VOLUME = {10},
YEAR = {2018},
NUMBER = {10},
ARTICLE-NUMBER = {1522},
ISSN = {2072-4292},
}

@article{verma2019transfer,
title = {Transfer learning approach to map urban slums using high and medium resolution satellite imagery},
journal = {Habitat International},
volume = {88},
pages = {101981},
year = {2019},
issn = {0197-3975},
author = {Verma, Deepank and Jana, Arnab and Ramamritham, Krithi},
}

@inproceedings{gramhansen2019mapping,
author = {Gram-Hansen, Bradley J. and Helber, Patrick and Varatharajan, Indhu and Azam, Faiza and Coca-Castro, Alejandro and Kopackova, Veronika and Bilinski, Piotr},
title = {Mapping informal settlements in developing countries using machine learning and low resolution multi-spectral data},
year = {2019},
booktitle = {proc. of the 2019 AAAI/ACM Conference on AI, Ethics, and Society},
pages = {361–368},
numpages = {8},
}

@misc{maiyababu2018slum,
  title={Slum segmentation and change detection: A deep learning approach},
  author={Maiya, Shishira R and Babu, Sudharshan Chandra},
  eprint={1811.07896},
  archivePrefix={arXiv},
  year={2018}
}

@misc{UNHABITAT_Data,
  author		= "UN-Habitat",
  title			= "Urban Indicators Database", 
  year			= "2025",
  howpublished		= "\url{https://data.unhabitat.org/}",
  note="Accessed: 2025-11-13"
}

@article{Wurm2019slum,
  author		= "Wurm, M. and Stark, T. and Zhu, X. X. and Weigand, M. and Taubenöck, H.",
  title			= "Semantic segmentation of slums in satellite images using transfer learning on fully convolutional neural networks",
  journal		= "ISPRS Journal of Photogrammetry and Remote Sensing",
  volume		= "150",
  pages			= "59--69",
  year			= "2019"
}

@misc{Nado2020EvaluatingPB,
  title={Evaluating Prediction-Time Batch Normalization for Robustness under Covariate Shift},
  author={Zachary Nado and Shreyas Padhy and D. Sculley and Alexander D'Amour and Balaji Lakshminarayanan and Jasper Snoek},
  eprint={2006.10963},
  archivePrefix={arXiv},
  year={2020},
}

@inproceedings{wang2021tent,
  title={Tent: Fully Test-Time Adaptation by Entropy Minimization},
  author={Wang, Dequan and Shelhamer, Evan and Liu, Shaoteng and Olshausen, Bruno and Darrell, Trevor},
  booktitle={proc. of the 2021 International Conference on Learning Representations},
  year={2021},
}

@inproceedings{wang2022continual,
  title={Continual Test-Time Domain Adaptation},
  author={Wang, Qin and Fink, Olga and Van Gool, Luc and Dai, Dengxin},
  booktitle={proc. of the 2022 Computer Vision and Pattern Recognition},
  year={2022}
}

@book{united2003challenge,
  title={The challenge of slums: global report on human settlements, 2003},
  author={UN-Habitat},
  year={2003},
  publisher={Routledge}
}

@article{taubenbock2014physical,
  title={The physical face of slums: A structural comparison of slums in Mumbai, India, based on remotely sensed data},
  author={Taubenb{\"o}ck, Hannes and Kraff, NJ},
  journal={Journal of Housing and the Built Environment},
  volume={29},
  number={1},
  pages={15--38},
  year={2014},
  publisher={Springer}
}

@article{kohli2012ontology,
  title={An ontology of slums for image-based classification},
  author={Kohli, Divyani and Sliuzas, Richard and Kerle, Norman and Stein, Alfred},
  journal={Computers, environment and urban systems},
  volume={36},
  number={2},
  pages={154--163},
  year={2012},
  publisher={Elsevier}
}

@article{simon2011situating,
  title={Situating slums: Discourse, scale and place},
  author={Simon, David},
  journal={City},
  volume={15},
  number={6},
  pages={674--685},
  year={2011},
  publisher={Taylor \& Francis}
}

@inproceedings{Yang2025Jurse,
  title={Towards consistent and robust slum detection 
using multi-year satellite data},
  author={Yang, Jeasurk and Lee, Sumin and Park, Sungwon and Ahn, Donghyun and Cha, Meeyoung},
  booktitle={proc. of IEEE Joint Urban Remote Sensing Event},
  pages={1--4},
  year={2025},
}

@article{kit2013automated,
  title={Automated detection of slum area change in Hyderabad, India using multitemporal satellite imagery},
  author={Kit, Oleksandr and L{\"u}deke, Matthias},
  journal={ISPRS journal of photogrammetry and remote sensing},
  volume={83},
  pages={130--137},
  year={2013},
  publisher={Elsevier}
}

@article{duque2017exploring,
  title={Exploring the potential of machine learning for automatic slum identification from VHR imagery},
  author={Duque, Juan C and Patino, Jorge E and Betancourt, Alejandro},
  journal={Remote Sensing},
  volume={9},
  number={9},
  pages={895},
  year={2017},
  publisher={MDPI}
}

@article{stark2020satellite,
  title={Satellite-based mapping of urban poverty with transfer-learned slum morphologies},
  author={Stark, Thomas and Wurm, Michael and Zhu, Xiao Xiang and Taubenb{\"o}ck, Hannes},
  journal={IEEE Journal of Selected Topics in Applied Earth Observations and Remote Sensing},
  volume={13},
  pages={5251--5263},
  year={2020},
  publisher={IEEE}
}

@inproceedings{yang2022st_plus,
    author = {Yang, Lihe and Zhuo, Wei and Qi, Lei and Shi, Yinghuan and Gao, Yang},
    year = {2022},
    title = {ST++: Make Self-training Work Better for Semi-supervised Semantic Segmentation},
    booktitle = {proc. of 2022 Computer Vision and Pattern Recognition},
}

@inproceedings{chen2023mod,
  title={Mod-squad: Designing mixtures of experts as modular multi-task learners},
  author={Chen, Zitian and Shen, Yikang and Ding, Mingyu and Chen, Zhenfang and Zhao, Hengshuang and Learned-Miller, Erik G and Gan, Chuang},
  booktitle={proc. of the 2023 Computer Vision and Pattern Recognition},
  pages={11828--11837},
  year={2023}
}

@inproceedings{lee2024becotta,
  title={BECoTTA: input-dependent online blending of experts for continual test-time adaptation},
  author={Lee, Daeun and Yoon, Jaehong and Hwang, Sung Ju},
  booktitle={proc. of the 41st International Conference on Machine Learning},
  pages={27072--27093},
  year={2024}
}

@article{yang2025ai,
  title={AI-based Ger detection reveals post-pandemic delay in informal housing progress in Mongolia},
  author={Yang, Jeasurk and Lee, Sumin and Park, Sungwon and Lee, Minjun and Cha, Meeyoung},
  journal={npj Urban Sustainability},
  volume={5},
  number={1},
  pages={78},
  year={2025},
  publisher={Nature Publishing Group UK London}
}

@inproceedings{park2021improving,
  title={Improving unsupervised image clustering with robust learning},
  author={Park, Sungwon and Han, Sungwon and Kim, Sundong and Kim, Danu and Park, Sungkyu and Hong, Seunghoon and Cha, Meeyoung},
  booktitle={proc. of the 2021 Computer Vision and Pattern Recognition},
  pages={12278--12287},
  year={2021}
}

@inproceedings{niu2023towards,
  title={Towards Stable Test-Time Adaptation in Dynamic Wild World},
  author={Niu, Shuaicheng and Wu, Jiaxiang and Zhang, Yifan and Wen, Zhiquan and Chen, Yaofo and Zhao, Peilin and Tan, Mingkui},
  booktitle = {proc. of the 2023 International Conference on Learning Representations},
  year = {2023}
}

@inproceedings{ahn2025generalizable,
  title={Generalizable disaster damage assessment via change detection with vision foundation model},
  author={Ahn, Kyeongjin and Han, Sungwon and Park, Sungwon and Kim, Jihee and Park, Sangyoon and Cha, Meeyoung},
  booktitle={proc. of the 39th AAAI Conference on Artificial Intelligence},
  volume={39},
  pages={27784--27792},
  year={2025}
}

@article{liang2025comprehensive,
  title={A comprehensive survey on test-time adaptation under distribution shifts},
  author={Liang, Jian and He, Ran and Tan, Tieniu},
  journal={International Journal of Computer Vision},
  volume={133},
  number={1},
  pages={31--64},
  year={2025},
  publisher={Springer}
}

@article{ahn2023human,
  title={A human-machine collaborative approach measures economic development using satellite imagery},
  author={Ahn, Donghyun and Yang, Jeasurk and Cha, Meeyoung and Yang, Hyunjoo and Kim, Jihee and Park, Sangyoon and Han, Sungwon and Lee, Eunji and Lee, Susang and Park, Sungwon},
  journal={Nature Communications},
  volume={14},
  number={1},
  pages={6811},
  year={2023},
  publisher={Nature Publishing Group UK London}
}

@inproceedings{xie2021segformer,
  author = {Xie, Enze and Wang, Wenhai and Yu, Zhiding and Anandkumar, Anima and Alvarez, Jose M. and Luo, Ping},
  title = {SegFormer: simple and efficient design for semantic segmentation with transformers},
  year = {2021},
  booktitle = {proc. of the 35th International Conference on Neural Information Processing Systems},
  articleno = {924},
  numpages = {14},
}

@inproceedings{liang2020shot,
  title = 	 {Do We Really Need to Access the Source Data? {S}ource Hypothesis Transfer for Unsupervised Domain Adaptation},
  author =       {Liang, Jian and Hu, Dapeng and Feng, Jiashi},
  booktitle = 	 {proc. of the 37th International Conference on Machine Learning},
  pages = 	 {6028--6039},
  year = 	 {2020},
  volume = 	 {119},
  publisher =    {PMLR},
}

@misc{slumdata_CapeTown,
  author		= "{{University of Edinburgh}}",
  title			= "Dwelling outline - Informal Settlements of Cape Town, 2018", 
  year			= "2020",
  publisher		= "School of Engineering. Infrastructure and Environment, University of Edinburgh",
  doi			= "10.7488/ds/2758"
}

@misc{slumdata_Mumbai,
  author		= "{{Slum Rehabilitation Authority}}",
  title			= "Georeferenced City Survey Plan of Mumbai City \& Suburban District Showing Area Boundaries of Slum Clusters", 
  year			= "2016",
  publisher		= "Slum Rehabilitation Authority, the Government of Maharashtra",
  howpublished		= "\url{https://sra.gov.in/page/innerpage/gis-mis-slum-data.php}"
}

@misc{slumdata_Karachi,
  author		= "{{Earth Observation for Sustainable Development}}",
  title			= "Service Operations Report - Karachi", 
  year			= "2018",
  publisher		= "European Space Agency",
  howpublished		= "\url{https://datacatalog.worldbank.org/search/dataset/0039832/Karachi--Pakistan----Informal-Settlements--ESA-EO4SD-Urban-}"
}

@misc{slumdata_Medellin,
  author		= "{{Alcaldia de Medellin}}",
  title			= "Informal Settlement Map", 
  year			= "2010",
  publisher		= "Empresa de Desarrollio Urbano",
  howpublished		= "\url{https://datacatalog.worldbank.org/search/dataset/0039832/Karachi--Pakistan----Informal-Settlements--ESA-EO4SD-Urban-}"
}

@misc{slumdata_RDJ,
  author		= "{{Prefeitura da Cidade do Rio de Janeiro}}",
  title			= "Limite de Favelas 2019", 
  year			= "2021",
  publisher		= "Prefeitura da Cidade do Rio de Janeiro",
  howpublished		= "\url{https://www.data.rio/datasets/limite-favelas-2019/explore}"
}

@misc{slumdata_AOI,
  author		= "Samper, J.",
  title			= "Atlas of Informality", 
  year			= "2025",
  howpublished = {\url{https://www.atlasofinformality.com/}},
  note = "Accessed: 2025-11-13"
}

@inproceedings{Radford2021CLIP,
  title={Learning Transferable Visual Models From Natural Language Supervision},
  author={Alec Radford and Jong Wook Kim and Chris Hallacy and Aditya Ramesh and Gabriel Goh and Sandhini Agarwal and Girish Sastry and Amanda Askell and Pamela Mishkin and Jack Clark and Gretchen Krueger and Ilya Sutskever},
  booktitle={proc. of the 38th International Conference on Machine Learning},
  year={2021},
}

@inproceedings{han2020lightweight,
  title={Lightweight and robust representation of economic scales from satellite imagery},
  author={Han, Sungwon and Ahn, Donghyun and Cha, Hyunji and Yang, Jeasurk and Park, Sungwon and Cha, Meeyoung},
  booktitle={proc. of the 34th AAAI Conference on Artificial Intelligence},
  volume={34},
  pages={428--436},
  year={2020}
}

@inproceedings{han2020learning,
  title={Learning to score economic development from satellite imagery},
  author={Han, Sungwon and Ahn, Donghyun and Park, Sungwon and Yang, Jeasurk and Lee, Susang and Kim, Jihee and Yang, Hyunjoo and Park, Sangyoon and Cha, Meeyoung},
  booktitle={proc. of the 26th ACM SIGKDD International Conference on Knowledge Discovery \& Data Mining},
  pages={2970--2979},
  year={2020}
}

@inproceedings{park2022learning,
  title={Learning economic indicators by aggregating multi-level geospatial information},
  author={Park, Sungwon and Han, Sungwon and Ahn, Donghyun and Kim, Jaeyeon and Yang, Jeasurk and Lee, Susang and Hong, Seunghoon and Kim, Jihee and Park, Sangyoon and Yang, Hyunjoo and others},
  booktitle={proc. of the 36th AAAI Conference on Artificial Intelligence},
  volume={36},
  pages={12053--12061},
  year={2022}
}

@ARTICLE{rehman2022temporary,
  author={Rehman, M. Fasi Ur and Aftab, Izza and Sultani, Waqas and Ali, Mohsen},
  journal={IEEE Geoscience and Remote Sensing Letters}, 
  title={Mapping Temporary Slums From Satellite Imagery Using a Semi-Supervised Approach}, 
  year={2022},
  volume={19},
  number={},
  pages={1-5},
}

@inproceedings{
zhang2024UVSAM, 
title={UV-SAM: Adapting Segment Anything Model for Urban Village Identification}, 
volume={38}, 
booktitle={proc. of the 38th AAAI Conference on Artificial Intelligence}, 
author={Zhang, Xin and Liu, Yu and Lin, Yuming and Liao, Qingmin and Li, Yong},
year={2024},
pages={22520--22528} }

@inproceedings{lin2024long,
  title     = {Long-term Detection and Monitory  of Chinese Urban Village Using Satellite Imagery},
  author    = {Lin, Yuming and Zhang, Xin and Liu, Yu  and Han, Zhenyu and Liao, Qingmin and Li, Yong},
  booktitle = {proc. of the 33rd International Joint Conferences on Artificial Intelligence},
  pages     = {7349--7357},
  year      = {2024},
}
